\documentclass[]{IEEEtran}
\usepackage{latexsym}
\usepackage{graphicx}
\usepackage{amsfonts,amssymb,amsmath}
\usepackage{hyperref}
\usepackage{siunitx}
\usepackage{color}
\usepackage{cite}
\usepackage[dvipsnames]{xcolor}
\usepackage[linesnumbered,ruled,vlined]{algorithm2e}

\def\BibTeX{{\rm B\kern-.05em{\sc i\kern-.025em b}\kern-.08em T\kern-.1667em\lower.7ex\hbox{E}\kern-.125emX}}

\begin{document}
\title{On Addressing Heterogeneity in Federated Learning for Autonomous Vehicles Connected to a Drone Orchestrator
} 

\author{Igor Donevski,\, Jimmy Jessen Nielsen, and Petar Popovski,
       
\thanks{I. Donevski, J. J. Nielsen, and P. Popovski are with Department of Electronic Systems, Aalborg University, Denmark (e-mail:\{igordonevski, jjn, petarp\}@es.aau.dk).}
}

\maketitle

\begin{abstract}
In this paper we envision a federated learning (FL) scenario in service of amending the performance of autonomous road vehicles, through a drone traffic monitor (DTM), that also acts as an orchestrator. Expecting non-IID data distribution, we focus on the issue of accelerating the learning of a particular class of critical object (CO), that may harm the nominal operation of an autonomous vehicle. This can be done through proper allocation of the wireless resources for addressing learner and data heterogeneity. 
Thus, we propose a reactive method for the allocation of wireless resources, that happens dynamically each FL round, and is based on each learner's contribution to the general model. In addition to this, we explore the use of static methods that remain constant across all rounds. Since we expect partial work from each learner, we use the FedProx FL algorithm, in the task of computer vision. For testing, we construct a non-IID data distribution of the MNIST and FMNIST datasets among four types of learners, in scenarios that represent the quickly changing environment. The results show that proactive measures are effective and versatile at improving system accuracy, and quickly learning the CO class when underrepresented in the network. Furthermore, the experiments show a tradeoff between FedProx intensity and resource allocation efforts. Nonetheless, a well adjusted FedProx local optimizer allows for an even better overall accuracy, particularly when using deeper neural network (NN) implementations. 

\end{abstract}
\begin{IEEEkeywords}
Federated Learning, FedProx, Contribution, Incentive, Staleness, Convergence, UAV
\end{IEEEkeywords}

\section{Introduction}
\label{introduction}

\begin{table*}[]
\begin{center}
    
\caption{Relevant symbols of variables, constants and functions.}
\begin{tabular}{lll}
\hline
Symbol &  Definition \\ \hline
\hline
$h_\text{MAX}()$ & Utility function that maximizes the number of computed epochs. \\
$h_\text{AAS}()$ & Utility function that minimizes the average anchored staleness.  \\
$h_\text{ACT}()$ & Utility function that maximizes based on the estimated contributions from each learner.  \\
$F()$ &  Local machine learning optimization function \\
$f()$ &  Global (network-wide) optimization function \\
$E()$ &  Model evaluation function \\
$\omega_{\text{g} \text{\textbackslash} \{k\} ,i}$ & Custom model aggregator that excludes the $k$ learner's model \\
$i$ & An integer indicating the FL cycle/round\\
$k$ & Learner index number\\
$K$ & Total number of leanrners in the MA \\
$\mathbf{T}_i$ & Vector representation of the epochs computed across all learners for round $i$\\
$\mathbf{G}_i$ & Vector representation of the contributions computed, for all learners, for round $i$\\
$\mathbf{S}_i$ & Vector representation of bandwidth allocated for each learner for round $i$\\
$G_{k,i}$ & Estimated contribution for learner $k$, at round $i$ \\
$\omega_{\text{g},i}$ & The global ML model weights for round $i$  \\
$\omega_{k,i}$ & The ML model weights produced at learner $k$ for round $i$   \\
$\tau_{k,i}$ & Epochs computed at learner $k$ for round $i$ \\
$B$ & The size of the batch computed at each epoch \\
$\mu$ & Proximal term intensity in the FedProx FL implementation \\
$f_k$ & Processing capability of learner $k$ in terms of epochs per millisecond \\
$W$ & Total bandwidth allocated for the system\\
$D$ & Total data transmitted in both directions to a single learner within a single round \\
$R_\text{avg}$ & Channel data rate in symbols per hertz  \\
$S_{k,i}$ & Bandwidth allocation coefficient for learner $k$ and round $i$ \\
$\alpha$ & Computation phase duration coefficient (in milliseconds) \\
$\beta$ & Communication phase duration coefficient (in milliseconds) \\
$S_{\text{min}}$ & The lower bound of the bandwidth allocation coefficient  \\
$S_{\text{max}}$ & The extreme bound of the bandwidth allocation coefficient   \\
\hline
\end{tabular}
\label{table:values}
\end{center}
\end{table*}

The adoption of ubiquitous Level-5 fully independent system autonomy in road vehicles (as per the SAE ranking system \cite{SAE}) is barred from progress due to the omnipresence of chaotic traffic in legacy traffic situations. Moreover, a 38\% share of prospective users are sceptical of the performance of the autonomous driving systems \cite{sceptics}. As such, lowering the number of negative outcome outliers in autonomous vehicle operation, particularly ones that lead to fatal incidents, can be addressed with an overabundance of statistically relevant data \cite{autonomous}. Thus, given the privacy requirements and the abundance of the data that is produced by road vehicles and/or unmanned aerial vehicles (UAVs) in the role of traffic monitors, the machine learning (ML) problem can be addressed by treating the participatory vehicles as learners in a federated learning (FL) network. 

In more detail, FL is an ML technique that distributes the learning across many learners. In this way, many separate models are aggregated in order to acquire one general model at server side \cite{googFL}. In FL, each learner does not have to send heaps of data to a common server for processing, but maintains the data privately. As such, the concept of FL is an extension of distributed ML with four important distinctions: (1) the training data distributions across devices can be non-IID; (2) not all devices have similar computational hardware; (3) FL scales for networks of just few devices to vast networks of millions; (4) FL can be engineered in a way in which privacy is conserved. Given the vast complexity of implementing FL in autonomous vehicular traffic, particularly related to the quickly changing environment, in this paper we focus on solving the issues of non-IID data learnt across several devices with unequal processing power.
We proceed with a review on relevant FL literature below.
\\
\subsection{State of the Art}
FL is an emergent field that has gained immense popularity in the last five years. From the relevant literature we highlight several works. \cite{TLisurvey} covers the state of the art regarding computational models, \cite{badsurvey} contains a clear understanding of the FL potential and its most prominent applications. \cite{Sur1} and \cite{Sur2} provide comprehensive coverage on the communications challenges for the novel edge computation, \cite{niknam2020federated} analyzes scenarios of FL where learners use wireless connectivity. Challenges and future directions of FL systems in the context of the future 6G systems is given in~\cite{6G}, while  \cite{industrial} elaborates upon the applications of FL on connected automated vehicles and collaborative robotics. \cite{resourceincentive} covers resource allocation and incentive mechanisms in FL implementations. Most of the works on FL concerning UAVs treat the devices as learners \cite{wsaad2,FLdrone1,FLdrone2}. This requires mounting heavy computational equipment on-board, and therefore it is an energy inefficient way of exploiting drones. In contrast, in our prior work~\cite{IgorStaleness} we have investigated techniques for reducing staleness when a UAV acts as an orchestrator by optimizing its flying trajectory.

There is also an interest in wireless resource allocation optimization for FL networks, as covered in the topics that follow. The work of \cite{jointlearn} proposes a detailed communications framework for resource allocation given complex wireless conditions and an FL implementation on IID data. This work has a strong contribution to the topic of convergence analysis of wireless implementations of FL with very detailed channel model. The work of \cite{wifed1} does a detailed convex analysis for distributed stochastic gradient decent (SGD) and optimizes the power allocation for minimizing FL convergence times. The work of \cite{wifed2} formulates FL over wireless network as an optimization problem and conducts numerical analysis given the subdivided optimization criteria. However, the aforementioned works perform their analysis on SGD which has been shown to suffer in the presence of non-IID data and unequal work times \cite{Fedprox}. The novel local subproblem that includes a proximal optimizer in \cite{Fedprox} achieves 22\% improvements in the presence of unequal work at each node.

The learning of both single task and multi task objectives in the presence of unequal learner contributions is a difficult challenge and has received a lot of attention, e.g. in the works of \cite{smith2017federated,li2019fair,agnostic}. This also leads to the question of analyzing contributions among many learners with vastly different hardware that is considered in works covering FL incentive  mechanisms, by \cite{resourceincentive,chen2020mechanism,fed1,fed2,incentives}. The incentive based FL implementations rely on estimating each learner's contribution and rewarding them for doing the work. Hence calculating appropriate rewards becomes a difficult challenge that also comes at the price of computation and communications as shown by \cite{rewards}. Such mechanisms are useful when orchestrating an FL where learners would collect strongly non-IID data and learn with vastly different processing capabilities.\\

\subsection{Drone Traffic Monitors as FL Orchestrators}
\begin{figure}[t]
\centering
\includegraphics[width=1\columnwidth]{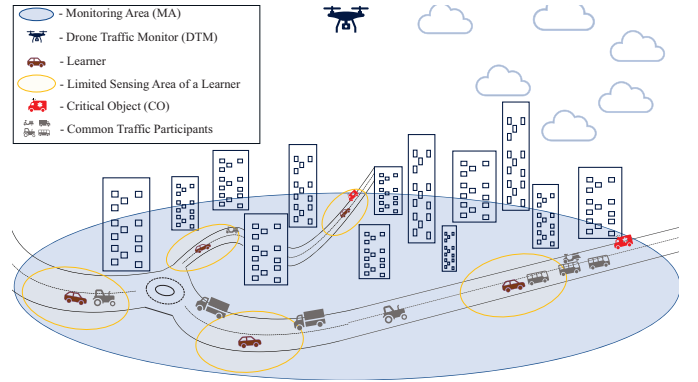}
\caption{Illustration of the DTM covered monitoring area, with five scattered learners.
}
\label{fig:toy}
\end{figure}

Unmanned aerial vehicles (UAVs) or drones could provide an essential aid to the vehicular communication networks by carrying wireless base stations (BSs). In combination with the 5G standardisation and the emerging 6G connectivity, drone-aided vehicular networks (DAVNs) \cite{davns} are capable of providing ultra reliable and low latency communications (URLLC) \cite{droneurllc,ppurllc} when issuing prioritized and timely alarms. In accord, most benefits of DAVNs come as consequence of the UAV's capability to establish line of sight (LOS) with very high probability \cite{tutorial}. The good LOS perspective also benefits visual surveillance, hence enabling UAVs to offer just-in-time warnings for critical objects (COs) that can endanger the nominal work of autonomous vehicles.
Though DAVNs expect many roles from the drone, we draw inspiration from UAVs in the role of drone traffic monitors (DTMs) that continuously improve and learn to perform timely and reliable detections of COs. To avoid requiring a plethora of drone-perspective camera footage of the traffic, we propose DTMs that take the role of a federated learning (FL) orchestrator, and autonomous vehicles participate as learners.

This FL architecture with a drone orchestrator, illustrated in Fig. \ref{fig:toy}, exploits the processing and sensing enabled vehicles contained in the monitoring area (MA) to participate both as learners and supervisors. The vehicle-learners receive the drone provided footage, and do the heavy computational work of ML training for the task of computer vision. This is possible since the vehicle-learners have robust sensing capabilities, and when they have the CO in view, can contribute to the learning process due to their secondary perspective \cite{chavdarova2018wildtrack} on the object, and their deeper knowledge of traffic classes. However, even when assuming perfect supervision by the learners, FL is not an easy feat since some knowledge can be obfuscated among omnipresent information and/or contained at computationally inferior straggler learners. In accord, we use a combination of state of the art FL implementation with a novel resource aware solution for balancing work times and learner contributions, which are described in the overview that follows.
\\
\subsection{Main Contributions}
In this paper, we provide a novel perspective on continuous DTM improvements through an FL implementation onto vehicle-learners. Moreover, we aim to provide a robust and adaptable resource allocation method for improved FL performance in the presence of chaotic, quickly changing, and most importantly imbalanced and non-IID data. Since both computational and data bias cannot be analytically extracted before sampling the ML model received from each learner, we assume heuristic measures such as maximizing the epochs computed, or equalizing the epochs computed across the learners. Moreover, the core contribution of this work is a dynamic resource allocation method based on each learner's past contributions. To provide full compatibility with heterogeneous learners and non-IID data, we employ these methods in combination with the FedProx algorithm. Finally, we developed an experimental analysis in which the performance is evaluated through its capability to learn an underrepresented class of the dataset, while also balancing overall system accuracy.

The paper organisation goes as follows. Section \ref{model} introduces the learning setup and the communications resource allocation setup. Section \ref{problem} defines the optimization problem and lists several static and reactive heuristic measures for improving the learning performance, and introduces the learner contribution calculations. This is followed by Section \ref{results} where the experimental setup and the results from the setup are presented. The final, Section \ref{conclusion} summarizes the outcomes and discusses future directions.

\section{System Model}

\label{model}
\begin{figure}[t]
\centering
\includegraphics[width=1\columnwidth]{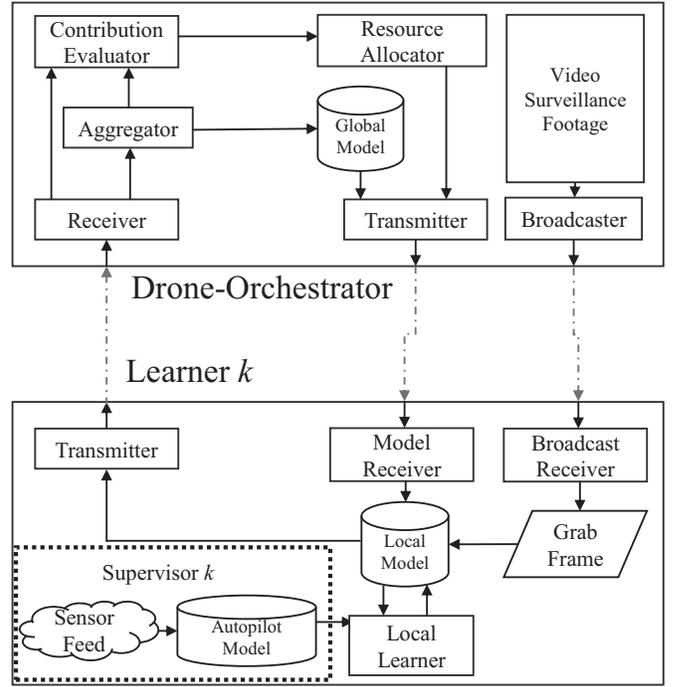}
\caption{System model illustration.
}
\label{fig:flow}
\end{figure}

The setup is depicted in Fig. \ref{fig:flow}, where we show the orchestrator block that sends and receives the models through wireless connections, while simultaneously broadcasts the unsupervised video surveillance footage at a constant data rate for all vehicles inside the MA. We assume that each vehicle acts as an ideal supervisor for the objects which are represented both in the broadcasted video and their sensor feed. 
Given some deadline of completion $T$, the learner needs to return its locally learnt model to the drone-orchestrator. After receiving the model, the orchestrator aggregates the $K$ models, after which it can also evaluate the contribution of each learner separately.  Each learner $k$ has a contribution, that the contribution estimator estimates to be $G_{k,i}$, for some FL cycle/round $i$. Finally, the orchestrator contains a resource allocator module that based on the aforementioned information can readjust the wireless resources for the next round, in a way that it improves the FL process.
\\
\subsection{Federated Learning}
The FL process starts when the orchestrator sends its weights to all $K$ learners, where each learner $ k \in \mathcal{K} =\{1,2,..,K\}$ is present in the MA. The goal of FL methods \cite{googFL} is to coordinate the optimization of a single global learning objective $\min_{\omega}f(\omega)$, where the function $f()$ is calculated across the whole network at each round $i$ as:
\begin{equation}
\label{eq:fedobjt}
f(\omega) = \sum_k^K p_k F_k(\omega) = \mathbb{E}[F_k(\omega)],
\end{equation}
where $\omega$ are the instantaneous value of the local model weights, $F_k(\omega)$ is the local optimization function at each node, $p_k \geq 0$ and $\sum_k p_k=1$ is the averaging weight when aggregating. 
In a single FL round $ i \in \mathbb{Z}^+$, a server, i.e. the DTM-orchestrator, has a global model with weights $\omega_{\text{g},i}$. On round $i$ each $k$-th learner receives the model and computes $\tau_{k,i}$ epochs of solving the local optimization function $F_k()$, with data batches of size $B$. Each batch represents a sample of items that have been sensed and collected from that learner's surroundings. The distributed training process produces a new set of weights $\omega_{k,i}$ at each $k$ that totals to $K$ different ML models. Hence, cycle $i$ concludes when all $\omega_{k,i}$ are aggregated to a signle set of weights $\omega_{\text{g},i+1}$, that serve as the collective model for the next iteration. The two most prominent approaches to solve the FL problem are Fedavg \cite{googFL} and Fedprox \cite{Fedprox} and differentiate mainly in the local optimization problem $F_k()$ at each device.

Using stochastic gradient descent (SGD) as a local solver $F_k()$, federated averaging (FedAvg) locks the amount of local epochs for each device to a fixed value. As such, each learner is fixed on computing the same $F_k()$ with the same learning rate of SGD for the same amount of epochs. For the successful operation of this system, it is essential to tune the optimization hyperparameters properly including the amount of epochs. The tradeoff in FedAvg becomes one of computation and communication since computing more local epochs reduces communication overhead at the expense of diversifying the local objectives as each system converges to a local optima given their portion of the non-IID data.

Due to the expected heterogeneity in the network of learners in the proposed FL implementation, we use the FedProx algorithm. The benefit of FedProx is that it can converge and provide good general models even under partial work and very dissimilar amounts of $\tau_{k,i}$. This is done by introducing a proximal term $ \lVert \omega - \omega_{\text{g},i} \rVert$ that alleviates the negative impact of the heterogeneity as:
\begin{equation}
\label{eq:fedprox}
F_k(\omega;\omega_{\text{g},i}) = L_k(\omega) + \frac{\mu}{2}  \lVert \omega - \omega_{\text{g},i} \rVert ,
\end{equation}
where $\omega$ is the instantaneous value of the local model weights at the local optimizer, $L_k(\omega)$ is a local cost function for the estimation losses, $\mu$ is a hyperparameter controlling the impact of the proximal term. The role of the proximal term here is that it prevents the local optimiser from straying far from the global model at round $i$. Moreover, we can control the local optimization problem to vary from a FedAvg ($\mu=0$) to FedProx ($\mu > 0$). We note that even when using Fedprox, too much local work causes the local optimizers to diverge from the global objective \cite{Fedprox}. Finally, using \eqref{eq:fedprox} for minimizing the local sub-problem $\min_{\omega}F_k(\omega;\omega_{\text{g},i})$ the FL converges to a solution even in the presence of heterogeneity and non-IID data distribution \cite{Fedprox}. Therefore, we use the FedProx algorithm to allow for full flexibility in data and processing heterogeneity, in combination with the resource allocation module that follows.
\\
\subsection{Allocation of Wireless Resources}
\begin{figure}[t]
\centering
\includegraphics[width=1\columnwidth]{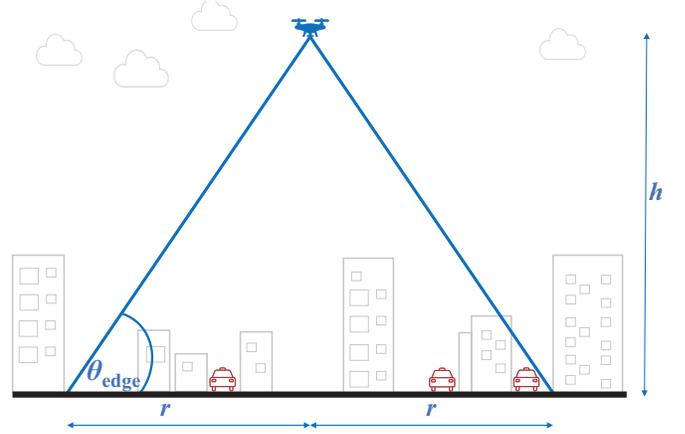}
\caption{Illustration of the drone position and geometry, in the communications setting.
}
\label{fig:dronpos}
\end{figure}

Though the work of \cite{jointlearn} covers a detailed cellular model for FL connectivity, drone provided connectivity is generally uniform and can be designed to be predominantly line of sight \cite{babu1}. As we illustrate in Fig. \ref{fig:dronpos} the drone height $h$ and the projected coverage on the ground with radius $r$ impact the elevation angle at the edge of the MA, $\theta_\text{edge}$. The steepness of the elevation can be derived from the evironmental parameters while also accounting for the directivity of the antenna mounted on the drone, as in \cite{igoreucnc}, and the service reliability that needs to be achieved \cite{igorwcnc}.

Since our goal of a DMT implementation is to improve the worst case performance of autonomous traffic, we also model the communications system through $\theta_\text{edge}$ as a worst case design parameter. $\theta_\text{edge}$ is decided upon deployment as it plays an important role of controlling the likelihood of establishing line of sight with the ground vehicles at the edge of the cell as in:
\begin{equation}
\label{eq:plos}
    \text{P}_\text{DLoS}  =\frac{1}{1+a\exp(-b(\theta_\text{edge}-a))},
\end{equation}
where $a$ and $b$ are constants defined by the propagation topology of the environment, as given by \cite{atg}. Through $\theta_\text{edge}$ in \eqref{eq:plos} a system designer controls not only the probability of detecting a CO but also the average quality of the communications channel at the edge of the MA as:
\begin{equation}
\label{eq:expected}
\Lambda  = L_\text{LoS} \cdot \text{P}_\text{DLoS} + L_\text{NLoS} \cdot ( 1 - \text{P}_\text{DLoS} ),
\end{equation}
where $L_\text{LoS}$ and $L_\text{NLoS}$ are the pathloss coefficients when LOS is established or lost, respectively. As such, we arrive to the average rate for the user located at the edge of the cell by:
\begin{equation}
\label{eq:rate}
R_\text{avg} = \log_2(1+\frac{P_\text{tx} }{ N \Lambda }) ,
\end{equation}
where $P_\text{tx}$ is the transmission power, and $N$ is the noise power. As FL model transmissions  usually take several seconds depending on the size of the model, we omit small scale fading as an impactful factor in the analysis and assume that the drone provided links are symmetrical in both directions and offer each learner $k$ a rate of $\frac{W}{K} \cdot R_\text{avg}$, where $W$ is the total bandwidth dedicated for the FL model passing. $W$ may be represented as discrete resource blocks or a band of spectrum that is left over after portioning part of it for the purpose of video broadcasting. Like this, $R_\text{avg}$ acts as a lower bound guarantee for the amount of time spent learning at each ground device. 

As the size of the processing batch is fixed to $B$, each device $k$ is tasked with an equal number of floating point operations (FLO) for each epoch, and computes $\tau_{k,i}$ epochs. However, for each learner $k$ we introduce a coefficient $f_k$ that represents the learners' computational power with regards to the model size, and is a unit of amount of epochs computed per unit time. Having full information on $f_k$ is generally trivial since it depends on the processing capabilities of the learner, which should be publicly available in the device specifications.

Given an equal bandwidth allocation to all devices, the total number of epochs is a linear function of $f_k$. This results in the following equation for $\tau_{k,i}$:
\begin{equation}
\label{eq:inittk}
\frac{\tau_{k,i}}{f_k} = T- \frac{KD}{WR_\text{avg}},
\end{equation}
where, $D$ is the total amount of data that needs to be sent in both directions within the deadline of $T$. We convert the problem to a step-wise nomenclature that gives the relationship between each learner, independent of the length of $T$ but as a relative inter-learner metric:
\begin{equation}
\label{eq:many}
\begin{split}
\tau_{k,i} - \tau_{l,i} &= T f_k- \frac{KD f_k}{WR_\text{avg}} - T f_l + \frac{KD f_l}{WR_\text{avg}},\\
\tau_{k,i} - \tau_{l,i} &= T (f_k - f_l) - (f_k-f_l) \frac{KD}{WR_\text{avg}},\\
\frac{\tau_{k,i} - \tau_{l,i}}{f_k - f_l} &= T - \frac{KD}{WR_\text{avg}}, 
\end{split}
\end{equation}
where $\,\, \forall k,l \in \mathcal{K}, l \neq k$. We then perform the substitution:
\begin{equation}
\begin{split}
\alpha &= \frac{\tau_{k,i} - \tau_{l,i}}{f_k - f_l}, \,\, \forall k,l \in \mathcal{K}, l \neq k, \\
T &= \alpha + \frac{KD}{WR_\text{avg}},
\end{split}
\end{equation}
where $\alpha$ is the nominal time reserved for learning, and it is directly influenced by the amount of FLOPs required to compute one epoch. This simplifies to:
\begin{equation}
\label{eq:many}
\begin{split}
\frac{\tau_{k,i}}{f_k}& = \alpha + \frac{KD}{WR_\text{avg}}- \frac{KD}{S_{k,i}WR_\text{avg}}, \\ 
\frac{\tau_{k,i}}{f_k}& = \alpha + \frac{KD}{WR_\text{avg}} (\frac{S_{k,i} - 1}{S_{k,i}}), \\ 
\end{split}
\end{equation}
where $S_{k,i} \geq 0$ and $\sum_k^K S_{k,i} = K$ is the bandwidth allocation for learner $k$ in round $i$, represented as the portion of the average spectrum $\frac{W}{K}$ occupied (i.e. $S_{k,i} = K$ is the full spectrum, and $S_{k,i} = 1$ is the average spectrum). We continue with the substitution: 
\begin{equation}
\frac{KD}{WR_\text{avg}} = \beta,
\end{equation}
where $\beta$ is the portion of time spent transmitting within one round.  As per $\beta$, it is obvious that it is much more important to investigate the ratio of data load on the channel instead of solely focusing on the achieved rate $R_\text{avg}$. Moreover, the time spent learning at each device becomes more significant the more we load the resources, in both number of learners and the size of the model. This results in the final representation of epochs computed for learner $k$ as a function of the bandwidth allocated to them:
\begin{equation}
\label{eq:tau}
\tau_{k,i} = {f_k}\alpha + {f_k}\beta (\frac{S_{k,i} - 1}{S_{k,i}}),
\end{equation}
Given a no-drop policy (each learner must complete at least one epoch $\tau \geq 1$), the lower bound on $S_{k,i}$ becomes:
\begin{equation}
\label{eq:lowbnd}
S_{\text{min}} = -\frac{\beta f_k}{1-\alpha f_k - \beta f_k},
\end{equation}
and the extreme upper bound of $S_{k,i}$ is therefore:
\begin{equation}
\label{eq:updnd}
S_{\text{max}} = K+\sum_l^{K-1} \frac{\beta f_l}{1-\alpha f_l - \beta f_l}, \,\, \forall l \in \mathcal{K}, l \neq k.
\end{equation}

The behaviour of the resource function for a single $\tau_{k,i}$ when adjusting $\beta$ and $S_{k,i}$ within the bounds of \eqref{eq:lowbnd} and \eqref{eq:updnd}, is:
\begin{equation}
\label{eq:bounded}
S_{\text{min}} \leq S_{k,i} \leq S_{\text{max}},
\end{equation}
The entire communications setup is reducible to the analysis of combinations of $\alpha$ and $\beta$, as both parameters directly determine the impact that resource allocation has on the system. Moreover, the parameter $\beta$ modifies the impact of resource allocation for each learner, where systems with high $\beta$ values stand to benefit the most, while low $\beta$ values indicate near instantaneous model transfers which cannot be influenced by modifying the bandwidth. On the other hand, $\alpha$ is a system design hyperparameter that indicates the amount of epochs computed within a single round, by an average learner, and it is fully customizable before or even during operation.

\section{Analysis}
\label{problem}
Our goal is to improve the learning of a particular class among the network of FL devices, that may represent a CO, without harming the overall accuracy of the system. Thus, each round $i$ we exploit our control over the wireless resources and optimize the bandwidth allocated to each device $S_{k,i}$. The vector representation of the bandwidth allocation for each round becomes $\mathbf{S}_i = (S_{1,i}, S_{2,i} ... S_{K,i})$. In the same way, the number of epochs computed in round $i$ and the contribution estimations are reformulated into vectors: $\mathbf{T}_i = (\tau_{1,i}, \tau_{2,i} ... \tau_{K,i})$ and $\mathbf{G}_i = (G_{1,i}, G_{2,i} ... G_{K,i})$ respectively, where $G_{k,i}$ is an estimate of the contribution of learner $k$ based of its learning performance in the past. Due to the rapidly changing environment around each learner, we cannot assume having information about the size or distribution of the data stored at each learner. Therefore, we can assume a function of utility from both aforementioned parameters $h_\text{X}(\mathbf{\tau}_i,\mathbf{G}_i)$, where X is a placeholder for the name of the approach. Given this function, the optimization problem of maximizing the utility $X$ can be defined as:
\begin{equation}
\label{eq:optprob}
\begin{split}
\max_{\mathbf{S}_i} \,\, & h_\text{X}(\mathbf{T}_i,\mathbf{G}_i), \\
& \sum_k^K S_{k,i} = K, \\
& \tau_{k,i} \in \mathbb{Z}^+,\\
& \eqref{eq:tau},\eqref{eq:lowbnd},\eqref{eq:updnd}, \eqref{eq:bounded}.
\end{split}
\end{equation}

Extracting the direct impact of $G_{k,i}$ and $\mathbf{\tau}_i$ onto the future accuracy of the model, and under non-IID data distribution, is non-trivial and hence requires that we form several heuristic functions for $h_\text{X}()$ to be tested on an experimental setup. Therefore we compare three different solutions for \eqref{eq:optprob} by swapping the utility function $h_\text{X}()$ with the ones named as $X \in \{ \text{MAX, AAS, ACT} \}$. The first two versions of the optimization problem (MAX and AAS) apply a static method that computes utility only as a function of the epochs that will be computed for that round for each learner. The third approach (ACT) is a novel reactive method, that extracts the utility of a learning round as a product of the estimated contribution by each learner and the epochs that will be computed by that learner. The details for each method follow below. \\

\subsection{Static Resource Allocation Measures}
The naive way of improving the convergence in a heterogeneous setting is maximizing the total amount of work done by all learners as in:
\begin{equation}
\label{eq:max}
h_\text{MAX}(\mathbf{T}_i,0) = \sum_k^K \tau_{k,i}.
\end{equation}
This optimization criteria maximizes the epochs computed across the whole network given the limited radio resources. Since \eqref{eq:max} implies asyncronous amount of work performed among the learners, it may not be considered as a potential maximization metric when using classical FedAvg implementations. However, since we use FedProx as a local optimizer, this is a sufficient naive solution that represents an exploitative behavior from the orchestrator. 

Furthermore, given the work on asynchronous FL and the issues of diverse computational hardware in the network \cite{async_fed_opt,async_solve} we identify maximum \emph{staleness} \cite{IgorStaleness} as an important criterion towards the precision of the model. We define this as the maximal difference between the fastest and slowest learner:
\begin{equation}
\label{eq:staleness}
s =  \max(|\tau_{k,i}-\tau_{l,i}|) \,\, \forall k,l \in \mathcal{K}, l \neq k.
\end{equation}
Nonetheless, minimizing staleness does not extract the full potential of our setup. Therefore, as in \cite{IgorStaleness} we convene $s$ and the average of the anticipated epochs to a more balanced heuristic metric, named Average Anchored Staleness (AAS) as an optimization metric:
\begin{equation}
\label{eq:aas}
h_\text{AAS}(\mathbf{T}_i,0) = \frac{1}{K}\sum_k^K \tau_{k,i} - s.
\end{equation}
AAS gives a good general overview that is data-agnostic, without the need to assume the impact of data at some particular learner and solely on spatial and computational performance. Like this, AAS provides a resource allocation objective function that serves an equally balanced amount of learning and \emph{staleness}.
\\

\subsection{Contribution Estimation for Reactive Resource Allocation}
In the case of DTMs, the considered vehicle supervisors/learners can find themselves in the presence of vastly different objects, and the data they sense changes constantly while they operate. Given the aforementioned, the contribution of each learner is hard to estimate especially in the presence of noisy samples. Hence, we assume that separating the important CO information ahead of time is impossible and only consider reactive approaches such as incentive mechanisms. To use incentive mechanisms we must assume that the validation dataset that is present at the orchestrator has equal representation of all classes. Hence, based on such validation data we can pass the weights $\omega$ through an evaluation function $E(\omega)$ which can be based on accuracy or loss evaluations of the model \textbf{(we choose accuracy)}. To calculate the contribution for each round $i$ we define:\begin{equation}
\label{eq:Gstep}
G_{k,i} = \frac{E(\omega_{\text{g},i})-E(\omega_{\text{g} \text{\textbackslash} \{k\} ,i})}{\sum_k^K |E(\omega_{\text{g},i})-E(\omega_{\text{g} \text{\textbackslash} \{k\} ,i})|},
\end{equation}
where $\omega_{\text{g} \text{\textbackslash} \{k\} ,i}$ is a model aggregator that constructs a new model that is an aggregate of all recieved models except the one of $k$. Hence the difference in accuracy between the fully aggregated model and the $\omega_{\text{g} \text{\textbackslash} \{k\} ,i}$ \cite{incentives} gives the added value (the uniqueness) of the learning done by learner $k$. Like this, the contribution estimator is capable of discovering the overall contribution from each learner for that round, without the capability of sampling for contributions on each detection class separately, or discern which object is underrepresented or is the CO. This is a central feature of our method, since we aim to improve CO learning without tailoring the solution to discern which class is the CO. 

We note that the $\omega_{\text{g} \text{\textbackslash} \{k\} ,i}$ function needs to be called for each learner in order to produce $K$ different contribution estimations. In addition to having to compute an additional parameter, there is one extra set of weights that needs to be aggregated for the calculation of $\omega_{\text{g} \text{\textbackslash} \{k\} ,i}$ for all other learners, thus making the complexity of the estimator scale as a square of the number of nodes in the system $K$. Even though the computational complexity of this technique can escalate in big FL implementations, in the architecture that we propose there should be several active learners inside the MA. Thus, even aside the limited computational power on the drone, the estimator module should not experience lengthy computational times. 

Following the first round, each device $k$ provides its model to the DTM-orchestrator. After which, the aggregator provides the first aggregate model weights $\omega_{\text{g},i}$.
The resource allocator module in the orchestrator receives the contributions for each of the participating learners and hence can decide to adjust the resources based on $G_{k,i}$. Since $G_{k,i}$ is an estimation of the contributions for the past round, the goal is to maximize the total contribution of the upcoming round by introducing the following optimization function:
\begin{equation}
\label{eq:alfareact}
h_\text{ACT}(\mathbf{T}_i,\mathbf{G}_i) = \sum_k^K \tau_{k,i} g({G_{k,i})},
\end{equation}
where $g()$ is a utility function that scales the contributions to match the impact of the number of computed epochs. Introducing a utility function is necessary to properly scale each learner's impact since $ -1 \leq G_{k,i} \leq 1 $ and $\tau_{k,i} \in \mathbb{Z}^+$. Since in an average scenario $\mathbb{E}[\tau_{k,i}] = \alpha\mathbb{E}[{f_k}]$, and $\mathbb{E}[{f_k}] =1 $ we scale our utility function as per the average epochs computed for that round as $g(G_{k,i}) =\alpha^{G_{k,i}}$. The bounds of the function become $ \frac{1}{\alpha} \leq g(G_{k,i}) \leq \alpha$, and the nominal non-contributive learners produce $g(0) = 1 $. Thus the heuristic exponential optimization function for the reactive solution can be calculated as the contribution corrected maximum epochs computed as in:
\begin{equation}
\label{eq:alfareact}
h_\text{ACT}(\mathbf{T}_i,\mathbf{G}_i) = \sum_k^K \tau_{k,i} \alpha^{G_{k,i}}.
\end{equation}
In the case of constantly equal contributions from all learners, the heuristic maximization criteria is reduced to the epoch maximization problem defined in \eqref{eq:max}. With $h_\text{ACT}$ defined as in \eqref{eq:alfareact} we maintain the problem within the bounds of mixed integer linear programming since the utility is applied only to $G_{k,i}$ that remains constant for the whole round $i$.

\section{Results}
\label{results}
\subsection{Experimental Setup}
For a set of learners that are scattered along the MA, our goal is to as closely as possible generate an experimental setup that simulates a realistic learner given the system model in Section \ref{model}. Since each learner has a very short amount of time to do the learning for the DTM, we approach the data as fleeting (stored very briefly) and concealed (cannot be known beforehand). Due to the complexity and the issues of reliably simulating the FL performance for full scale traffic footage, we test the performance of the proposed methods through simple and easily accessible computer vision datasets.
Each testing scenario was built using either the MNIST dataset \cite{mnist} of handwritten digits, or the FMNIST \cite{fminst} dataset consisting of 10 different grayscale icons of fashion accessories. 

As we expect that each vehicle contains strongly non-IID data we create a custom data distribution among $K=7$  learners as shown in Table \ref{table:learners}. In addition, the processing power for computing a certain amount of epochs per millisecond $f_k$ for each learner, is distributed as: two standard vehicles ($f_k$ = 1), two premium vehicles ($f_k$ = 1.3), and two budget vehicles ($f_k$ = 0.7); with the addition of one straggler that contains an older technology ($f_k$ = 0.15). At each epoch the learner samples a single batch of $B=16$ randomly selected values from the stored data (as per Table \ref{table:learners}). Like this, the training data changes constantly, to mimic the changing environment of the vehicular scenario. This makes this FL testing scenario unique in that the number of epochs computed also reflects the amount of data sampled from the environment.

\begin{table}[]
\begin{center}
    
\caption{The non-IID distribution of data among learners, and their computational coefficients $f_k$.}
\begin{tabular}{lll}
\hline
Learner & Classes Stored (out of 0-9) &  $f_k$ \\ \hline
\hline
$1$ & 3, 4, 5, 6 & 0.15 \\
$2$ & 0, 1, 2, 3, 4 & 0.7 \\
$3$ & 4, 6, 7, 8, 9 & 1.0 \\
$4$ & 0, 1, 2, 6, 7, 8, 9 & 1.3 \\
$5$ & 0, 1, 2, 7, 8, 9 & 1.3 \\
$6$ & 0, 1, 2, 7, 8, 9 & 1.0 \\
$7$ & 3, 4, 6 & 0.7 \\
\hline
\end{tabular}
\label{table:learners}
\end{center}
\end{table}
In the described setting, the class-number $5$ (6th class counting from zero) assumes the role of a CO. In addition to the CO, class-number $3$ is another non-CO class that is not too common and appears at only 3 learners. This is an over-exaggerated situation of having the CO data hidden at one node that is also a straggler. We expect this to be a realistic reflection of data in drone orchestrated FLs as nodes carry only a small amount of supervisory data for each class due to the fact that they stumble upon important objects randomly.

For detection, we implement a small convolutional neural network (CNN), common for the global and local models implemented in python tensorflow \cite{tensorflow}. In more detail, the CNN has only one 3x3 layer of 64 channels using the rectifier linear unit (ReLU), that goes to a 2x2 polling layer. A dense, fully connected neural network (NN) layer of 64 ReLU activated neurons receives the polled outputs of the convolutional layer, which is then fully connected to a NN layer of 10 soft-max activated neurons, one for each of the 10 categories of the NIST dataset. The local optimizer at each learner is given by the FedProx calculation in Eq. \eqref{eq:fedprox}, where the cost function $L_k()$ is a categorical cross-entropy loss function, and the learning rate performed well when fixed to $\gamma=0.1$. The communication phase coefficient was considered in milliseconds and chosen as $\beta=100$ considering our CNN model with a size of 2.5Mb that needs to be transmitted to all 7 learners, over a single $W=80$MHz 802.11ax channel. Finally, in the reference frame of milliseconds, the cycle duration coefficient was set to $\alpha=100$ in favor of allowing for higher flexibility when scaling the bandwidth allocation. \\

\subsection{MNIST Testing}

\begin{figure*}[th!]
\centering
\includegraphics[width=1\linewidth]{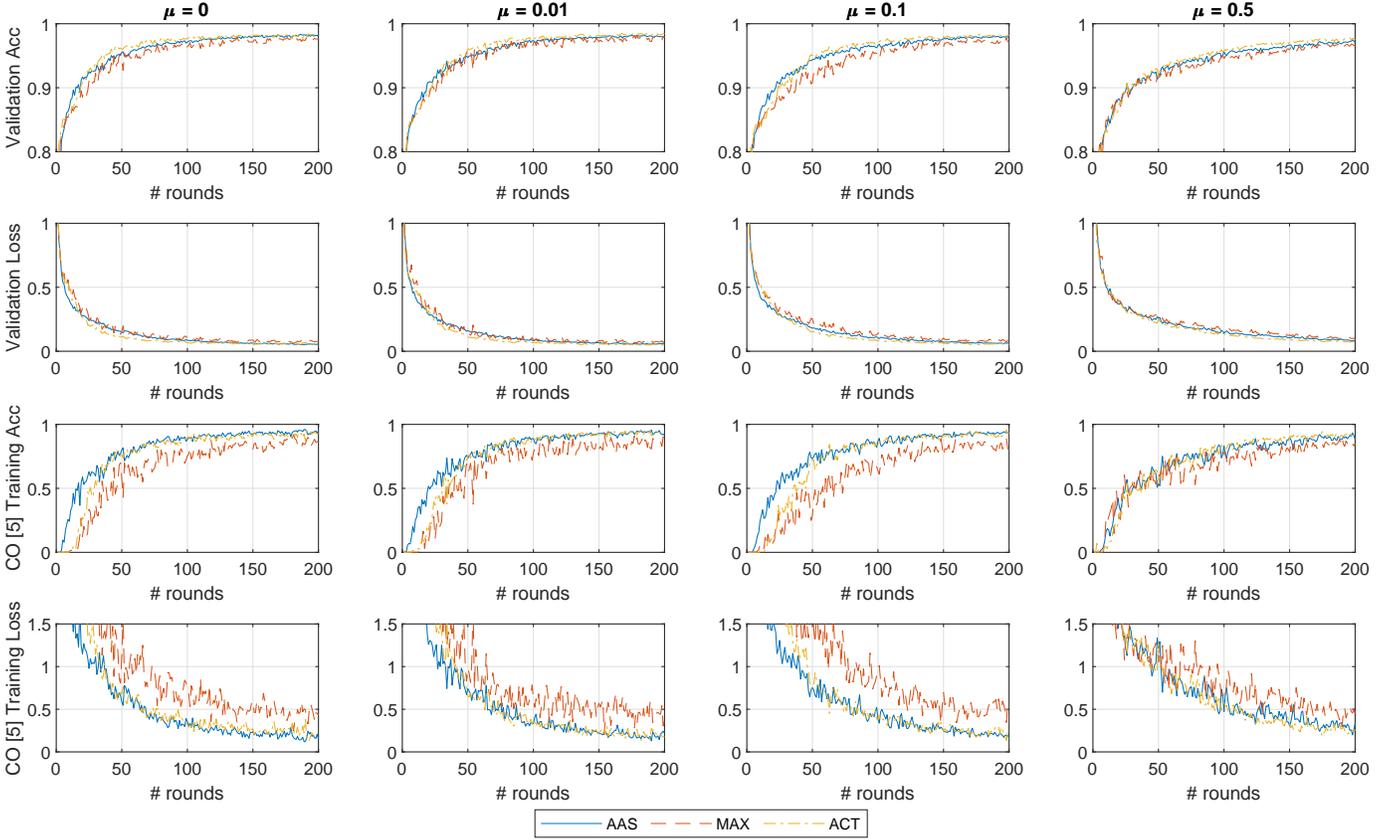}
\caption{General and CO-specific accuracy and loss results obtained when testing all three methods in combination with FedProx using the MNIST dataset.
}
\label{fig:mnist}
\end{figure*}

We proceed with the testing of all three approaches for five different values of the proximal importance hyperparameter $\mu \in \{ 0, 0.01, 0.1, 0.5\} $, as guided by the recommended values in \cite{Fedprox}. $\mu$ values larger than 0.5 failed to produce productive results and only harmed the convergence outlook. The testing lasts for 200 rounds on the aforementioned CNN model. Aside the three shown FL implementations, we also implement a classical ML with only one learner that contains all the data. We do this to extract the performance ceiling of the NN approach, which is 98\% for the validation accuracy and 0.0602 validation loss paired with training accuracy of 98.85\% and training loss 0.0423. 
\begin{figure}[t]
\centering
\includegraphics[width=1\columnwidth]{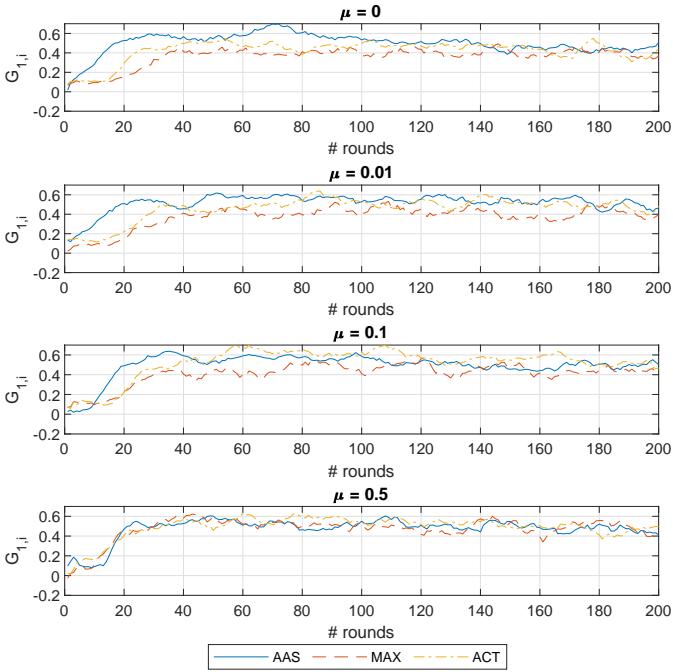}
\caption{Contribution evolution for learner $k=1$ in the case of using the MNIST dataset.
}
\label{fig:mnistcontrib}
\end{figure}

In Fig. \ref{fig:mnist} we can notice a limited impact of changing the $\mu$ parameter of FedProx, most likely due to the small amount of learners and not as significant straggler impact. This is expected given that \cite{Fedprox} claim strong superiority over FedAvg in the cases of very large portions of stragglers. Interestingly, $\mu$ does not have a strong positive impact on the learning performance even in the case of MAX, and therefore, a system designer would most likely introduce a weak proximal term of $\mu=0.01$. Additionally, using the ACT approach provides superior convergence, and in combination with $\mu=0.01$ achieves the best overall accuracy. In addition to this, the ACT and $\mu=0.01$ combination also keeps up with the performance of AAS with regards to the CO class after the first several rounds of convergence. 

To better investigate the behavior of the ACT approach we illustrate the evolution of the estimated contributions for learner $k=1$ in Fig. \ref{fig:mnistcontrib}, where $G_{1,i}$ is based on the performance of the learner estimated from the previous learning round as in Eq.~\eqref{eq:Gstep}. The overall conclusion here is that we achieve CO learning without tailoring the solution to discern which class is the CO. This is possible as the calculation of $G_{k,i}$ is focused around the uniqueness of the dataset at each learner.  Here we can notice that increasing the strength of the proximal parameter through setting higher $\mu$ values equalizes the contributions between all three methods, particularly in the first 40 rounds. Moreover, when $\mu = 0.5$ the contributions are stabilized and vary very little once the initial phase of 40 rounds. 

Most notably, the accuracy of AAS suffers significantly when $\mu=0.5$ which results in a performance that is equally matched to the MAX approach when detecting the CO. It is thus evident that a strong FedProx implementation harms total system accuracy, and above all, diminishes the impact of the using resource allocation.  Finally, we conclude that the task of learning MNIST is too simplistic for our assumed scenario of traffic monitoring, and thus we continue with testing the FMNIST dataset in the following subsection.
\\
\subsection{FMNIST Testing}
\begin{figure*}[t]
\centering
\includegraphics[width=1\linewidth]{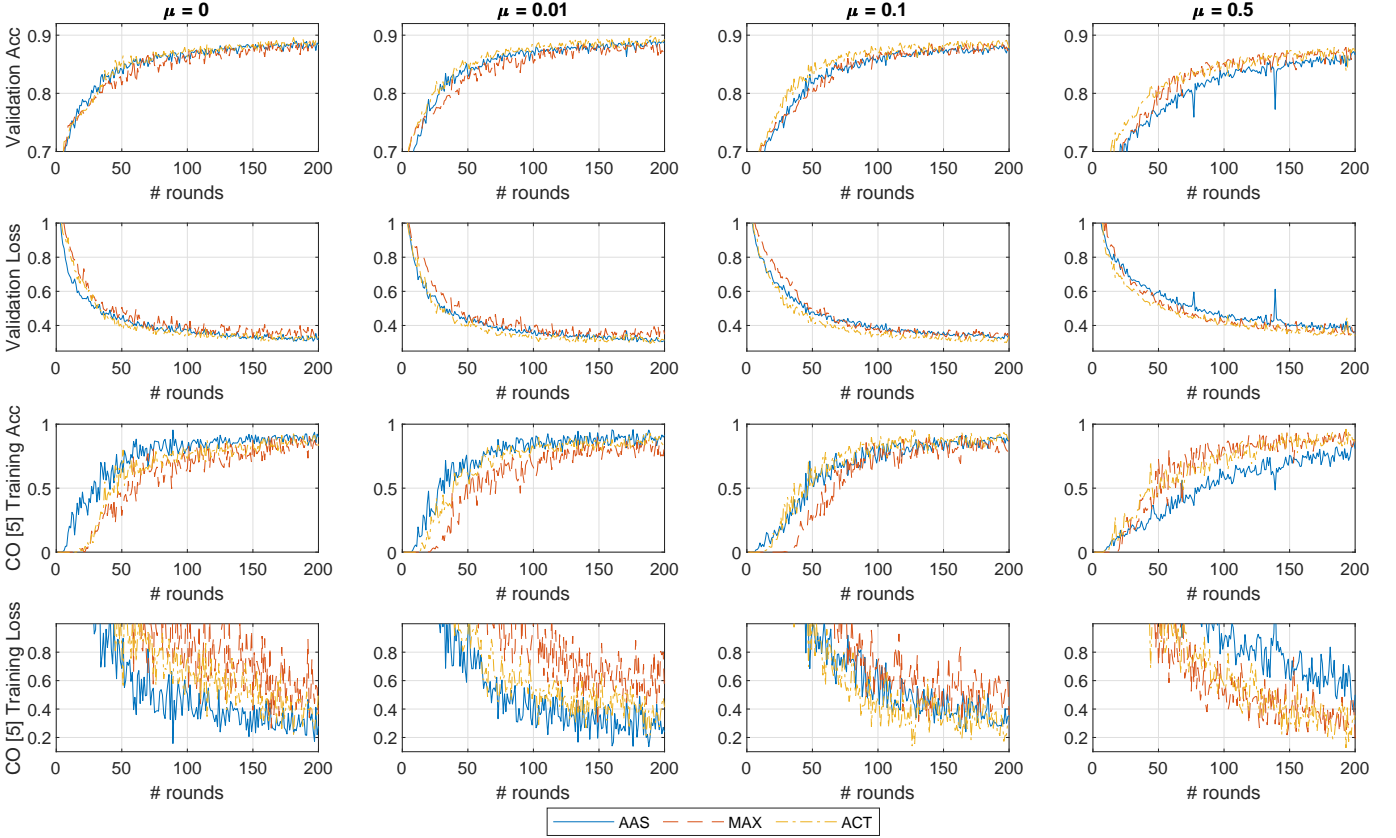}
\caption{General and CO-specific accuracy and loss results obtained when testing all three methods in combination with FedProx using the FMNIST dataset.
}
\label{fig:fmnist}
\end{figure*}

Since modeling common tasks of computer vision on MNIST is a very easy task we repeat the test on the FMNIST dataset. This dataset consists of 10 classes of fashion accessories in equal distribution as the MNIST dataset (a training set of 60\,000 examples and a test set of 10\,000 examples) and as in the case of MNIST consists of 28x28 grayscale images. The dataset classes are: (0) T-shirt/top, (1) Trouser, (2) Pullover, (3) Dress, (4) Coat, (5) Sandal, (6) Shirt, (7) Sneaker, (8) Bag, and (9) Ankle boot; where each item is taken from a fashion article posted on Zalando. Compared to the number MNIST, in FMNIST the intensity of each voxel plays a much bigger role and is scattered across larger parts of the image. We consider the FMNIST dataset as a computer vision task that sufficiently replicates the problem of detecting 10 different types of vehicles, in a much more simplistic context that is furthermore easily replicable.  

In Fig. \ref{fig:fmnist}, we show the learning performance in the same setting and $\mu \in \{ 0, 0.01, 0.1, 0.5\}$, across 200 rounds of training. It is most obvious that the overall accuracy has dropped quite a lot from the 98\% in the MNIST case to 88\% in the best case scenario of ACT with $\mu = 0.01$ for the FMNIST. Most notably the largest difference is that the increased difficulty of the learning problem introduces a lot more noise in the learning process, particularly for the CO class. Due to this, when using no FedProx ($\mu=0$) AAS does a good job at accelerating the learning process in the first 20 rounds until it is overtaken by ACT. Even though the combination of ACT with $\mu = 0.01$ shows the best overall accuracy on the validation data, the accuracy of detecting the CO class with ACT never truly reaches the performance of AAS. 

Finally, we conclude that even though $\mu = 0.1$ and $\mu = 0.5$ were eligible in the MNIST run, the overall increased complexity of FMNIST harms the accuracy outlook in both, but with the most severe impact on AAS. This experimental run therefore inspired us to investigate the issue of underfitting, and we proceed with testing FMNIST performance with a deeper model.
\\
\subsection{Deeper FMNIST Testing}
\begin{figure*}[th!]
\centering
\includegraphics[width=1\linewidth]{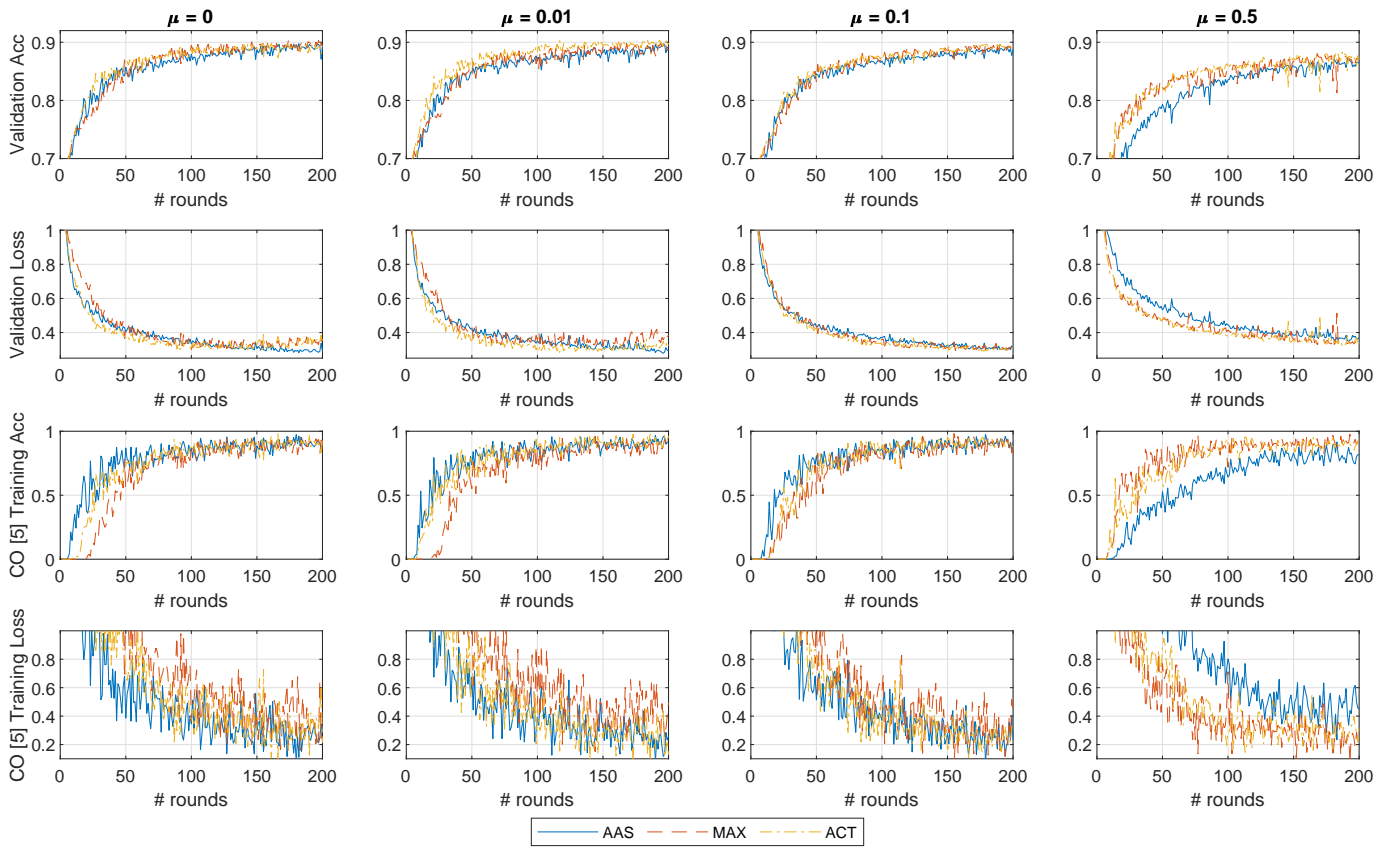}
\caption{General and CO-specific accuracy and loss results obtained when testing, with an extra CNN layer, all three methods in combination with FedProx using the FMNIST dataset.
}
\label{fig:lfmist}
\end{figure*}
\begin{figure}[th!]
\centering
\includegraphics[width=1\columnwidth]{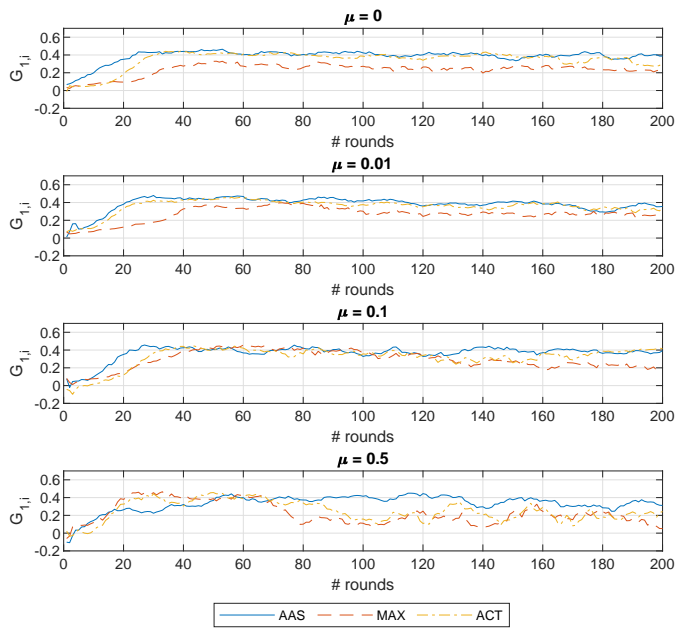}
\caption{Contribution evolution for learner $k=1$ in the case of using the FMNIST dataset with an extra CNN layer.
}
\label{fig:lltcontrib}
\end{figure}
In this testing scenario we expand the small convolutional neural network by adding another 3x3 layer of 64 channels using ReLU activators as a first layer. In Fig. \ref{fig:lfmist} we show the outcomes of the testing, where the overall accuracy of the system has been improved to 90\%. However, the larger model acted as an equalizer across all three approaches and in the case of $\mu=0$ generally gave equal performance both in convergence time and overall accuracy. It is important to also look at the validation loss following the round $i=150$ as it starts to diverge for both ACT and MAX approaches. This did not directly map into the accuracy of the detection, but nonetheless is a first sign of possible overfitting and eventual divergence.

With the deep model, this effect is diminished for the case of ACT with $\mu=0.01$, and manages to reach the best convergence time along with overall accuracy from all tested implementations. This accuracy is also paired with improved detection of the CO that exactly matches the AAS approach. As such the ACT with $\mu=0.01$ is both the best overall learning solution, but also the best CO detector.

It is also interesting to notice that the MAX approach does well with overall accuracy, particularly when compared to the inferior performance in the previous testing sets. Nonetheless, MAX is still inferior to both other approaches when it comes to detecting the CO class. Finally, we focus on the results on $\mu=0.5$. When the proximal term has such a strong impact on the learning, all three approaches show inferior overall performance by 4-5 percentage points with regards to the best performing  $\mu=0.01$. However, it is interesting to see that the impact is by far most severe on the AAS approach, even reducing the CO detection performance. Additionally, MAX gives the best result when it comes to learning the CO behavior for $\mu=0.5$. Opposed to the behavior back in the MNIST testing, here AAS suffers from the increased complexity of the task, and in combination with a very strong proximal term reduces the overall learning of detection. This makes it is easy to conclude that a strong proximal term reduces the effect of resource allocation efforts.

We seek to discover the culprit for the inferiority of AAS in CO discovery when $\mu=0.5$ by plotting the contributions of learner $k=1$ in Fig. \ref{fig:lltcontrib}. Looking at the contribution evolution in case $\mu=0.5$ we extrapolate that AAS aims to keep the learner relevant while the reduced amount of learning across the whole network harms the potential contribution of all other nodes. This leads us to the final conclusion of this experiment which is that the ACT based approach is extremely versatile in providing good CO detection and accuracy even in the cases of $\mu=0$, a properly assigned $\mu$, and overly restricted FedProx implementation.
\\
\subsection{Testing Fleeting FMNIST}

The final test with the experimental setup is constructed such that we introduce stress in the learning process by introducing temporary losses in the supervision process. This is done by introducing a likelihood that a learner $k$ loses access to a detection class. This would be representative of a learner losing LOS of the object was able to supervise, and is therefore modelled as a two state markov model (such as the Gilbert Elliot \cite{7881090}) that has a good and a bad state. Hence each supervisor has $p=0.9$ chance to maintain supervision for that class (stay in the good state), and $1-p=0.1$ probability to lose supervision capability (and move to the bad state). If the vehicle loses supervision capabilities for that class, it has $r=0.5$ probability to maintain that state (remain in the bad state) or $1-r=0.5$ probability to regain supervision of that class. The values for the state transitions in the Gilbert-Elliot model were chosen with the experimental setup in mind so that not too much data is lost with regards to the previous testing setups. These testing parameters were provisioned arbitrarily, because higher values would make the learning process very lengthy imposing unrealistic testing times for our experiment, but still provide a lot of stress to the learning system.

\begin{figure}[t]
\centering
\includegraphics[width=1\columnwidth]{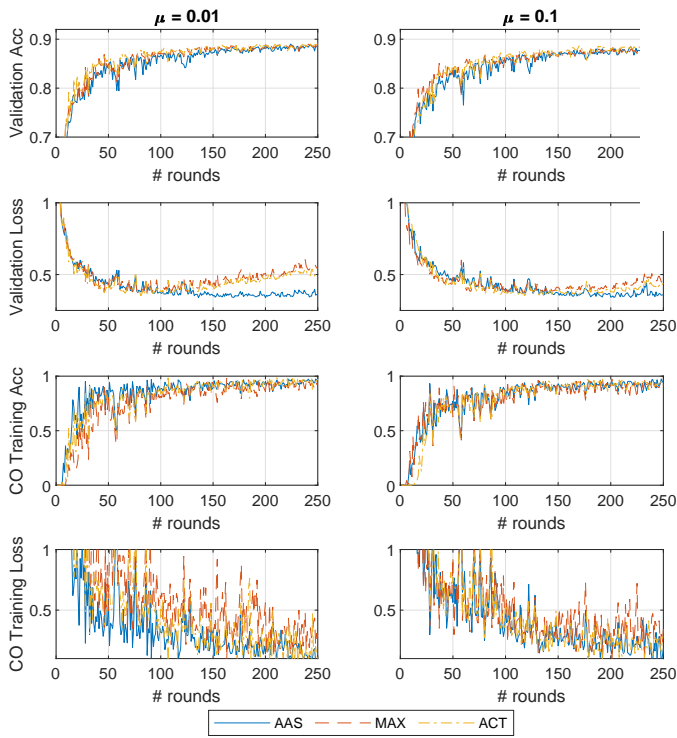}
\caption{General and CO-specific accuracy and loss results obtained when testing, with an extra CNN layer, all three methods in combination with FedProx using the FMNIST in the case of fleeting data.
}
\label{fig:fleet}
\end{figure}

Hence, to compensate for the smaller dataset, we let the simulations run for 250 rounds, and focus only on $\mu \in \{ 0.01, 0.1 \}$. The fleeting data is provided from the same seed and the Gilbert Elliot model starts from the good state for every possible detection combination. In Fig. \ref{fig:fleet} we show the performance of all approaches on the aforementioned setup. Comparing this to the previous testing setup, we notice that the overall accuracy dropped by 1 percentage point for $\mu=0.01$ and 2 percentage points when $\mu=0.1$ due to the increased stress in the learning process. It is also apparent that both ACT and MAX show signs of overfitting -- the diverging lines in the validation loss -- which is improved when using $\mu=0.1$, at the cost of reducing the overall system accuracy by an additional 1 percentage point. 

\begin{figure}[t]
\centering
\includegraphics[width=1\columnwidth]{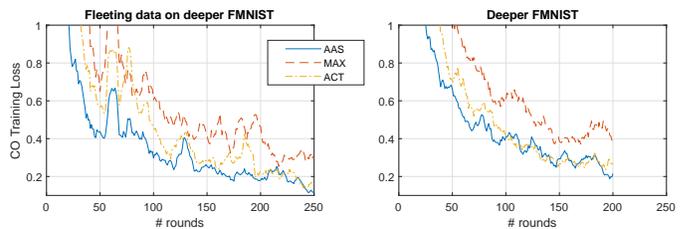}
\caption{10-point moving average of CO training loss for $\mu=0.01$ of the fleeting data versus normal data sampling in the deeper FMNIST testcase.
}
\label{fig:fleetcomp}
\end{figure}

Focusing on $\mu=0.01$, all methods achieve nearly the same overall accuracy, since the learning of the computer vision task is bottlenecked by the presence of the data. However, AAS is superior in CO detection and it shows slightly inferior convergence time for overall accuracy (i.e. around the 50 round mark). In addition to this, AAS is the most data sensitive approach and experiences the largest overall accuracy dips in situations where many detection classes are in the bad state (such as around the 55th round and the 127th round). Finally, to better observe the noisy training data, we plot a 10-point moving average in Fig.~\ref{fig:fleetcomp}. Here we notice the in the common training scenario AAS and ACT perform rather equally when learning hidden information. However, in the presence of fleeting data, the ACT performance becomes very noisy and becomse slightly inferior than AAS with regards to CO learning performance. Nonetheless, as already mentioned, this CO learning performance of the AAS approach comes at a slight cost of general detection performance, in both fleeting and normal setting. 
\subsection{Key Takeaways}
We condense several takeaways that were derived from all four experimental runs. The initial and most important conclusion is that the concepts of resource allocation and FedProx are at odds in the case of FL implementations. In more detail, the goal of FedProx is to reduce the impact of each learner individually while resource allocation methods strive to improve the overall performance by exploiting or compensating the heterogeneity of the system. Hence the impact of resource allocation methods is diminished when strengthening the role of the proximal term. Nonetheless, in the many tests a safe balance between both $\mu$ and resource allocation ensure good learning behavior. As such, we recommend that all future works consider perturbed gradient descent implementations, such as FedProx, when dealing with non-IID data in heterogeneous FL. 

Additionally, in the initial testing of our setup we noticed that testing on MNIST is not sufficient to provide reasonable results for the implementations, due to how trivial the task of recognising digits is. Moreover, FL implementations, such as the proposed drone implementation, are based in the distributed learning of complex tasks and require deeper NN models. In such cases, it was evident that increasing the total amount of computed epochs benefits the convergence time of the system with potentially harmful effects in CO detection accuracy. Moreover, deeper model implementations did not behave well under strong proximal terms.

As a consequence to this, learning hidden data can be addressed by equalizing the contributions by using AAS or by introducing strong proximal terms. However, the strong proximal terms have potential to slow down the convergence time for all nodes. Hence, the safest implementation to achieving the best combination of convergence time, overall accuracy and CO learning rate is using the ACT approach with a weak proximal term.

Finally, in a case where the data is fleeting, using a $\mu>0$ was crucial to reach stable learning performance. In this setting, the low availability of data acted as a lower bound for all learning implementations, but most importantly harms the convergence time performance of AAS. This is understandable since AAS was the approach that cumulatively computed the least amount of epochs at each round. On the other hand, the ACT approach maintained superior performance to both static approaches by maintaining good CO detection performance and great convergence times.

Finally, we extrapolate that defining a proper $\mu$ is cardinal. However, the hyperparameter needs to be defined ahead of the deployment of the system. As such, since we would not have access to the training data, the feasibility of implementing AAS is uncertain especially for situations where the presence of data changes quickly. This gives another strong motivation for using reactive measures based on contributions and incentive calculations, such as ACT.
\section{Conclusion}
\label{conclusion}
In this paper we investigated the learning process in a novel Federated Learning (FL) architecture, where a DTM acts as an orchestrator and traffic participants act as supervisors on its model. Such an implementation expects impairments on the learning process due to unbalanced and non-IID data scattered across heterogeneous learners that have variable computational equipment. We therefore test the ability of two static methods (AAS and MAX), and one incentive based reactive (ACT) resource allocation method to improve the speed of learning CO classes and maintaining good overall model accuracy. The validity of the methods was tested with an experimental FL implementation that uses the novel FedProx algorithm to learn from the MNIST and FMNIST datasets. The testing was conduced across combinations of different FedProx strength, CNN model depth, and fleeting data. 
From the testing we conclude that both reactive (ACT) resource allocation and FedProx are essential to securing model accuracy. In more detail, due to the inability to anticipate the distribution of the data across the learners, the use of ACT ensures proper operation of the FL implementation. In accord, the combination of properly set FedProx with an ACT implementation provided faster convergence times, better accuracy, but most importantly it matched the AAS method in learning to recognize the CO. Such behavior was consistent across most runs given the varying task complexity, model size, and data presence. The goal of future works would be to look into more advanced proactive approaches, especially for the presence of imperfect data supervision.

\section*{Conflict of Interest Statement}
The authors declare that the research was conducted in the absence of any commercial or financial relationships that could be construed as a potential conflict of interest.
\section*{Author Contributions}
ID: investigation, writing; JJN and PP: writing, review, editing, resources, funding acquisition, supervision, and project administration.
\section*{Funding}
The work was supported by the European Union's research and innovation programme under the Marie Sklodowska-Curie grant agreement No. 812991 ''PAINLESS'' within the Horizon 2020 Program.

\bibliographystyle{IEEEtran}
\bibliography{./bibliography.bib}

\end{document}